\documentclass[letterpaper, 10 pt, conference]{ieeeconf}  
\usepackage{caption}
\usepackage{subcaption}

\usepackage{mathtools}
\usepackage{amssymb}
\usepackage{balance}
\usepackage{dirtytalk}

\usepackage[algo2e,linesnumbered,ruled,vlined]{algorithm2e}

\usepackage{hyperref}

\usepackage{todonotes}

\newcommand*\rot{\rotatebox{90}}  

\IEEEoverridecommandlockouts                              

\DeclareCaptionLabelSeparator{periodspace}{.\quad}
\title{\LARGE \bf
One-Shot Open-Set Skeleton-Based Action Recognition
}

\author{Stefano Berti, Andrea Rosasco, Michele Colledanchise, and Lorenzo Natale  
\thanks{The authors are with the  Humanoids Sensing and Perception, Istituto Italiano di Tecnologia. Genoa, Italy.
e-mail: {\tt{ stefano.berti@iit.it}} } }

\begin{document}

\maketitle
\thispagestyle{empty}
\pagestyle{empty}

\begin{abstract}
Action recognition is a fundamental capability for humanoid robots to interact and cooperate with humans. This application requires the action recognition system to be designed so that new actions can be easily added, while ``unknown'' actions are identified and ignored.
In recent years, deep-learning approaches represented the principal solution to the Action Recognition problem. 
However, most models often require a large dataset of manually-labeled samples. 
In this work we target One-Shot deep-learning models, because they can deal with just a single instance for class.
Unfortunately, One-Shot models assume that, at inference time, the action to recognize falls into the support set and they fail when the action lies outside the support set. Few-Shot Open-Set Recognition (FSOSR) solutions attempt to address that flaw, but current solutions consider only static images and not sequences of images. Static images remain insufficient to discriminate actions such as sitting-down and standing-up.

In this paper we propose a novel model that addresses the FSOSR problem with a One-Shot model that is augmented with a discriminator that rejects ``unknown'' actions. 
This model is useful for applications in humanoid robotics, because it allows to easily add new classes and determine whether an input sequence is among the ones that are known to the system.
We show how to train the whole model in an end-to-end fashion and we perform quantitative and qualitative analyses. Finally, we provide real-world examples.
\end{abstract}

\section{Introduction}
Action Recognition (AR) represents a fundamental task for Human-Robot Interaction \cite{rodomagoulakis2016multimodal,akkaladevi2015action,roitberg2014human}, video surveillance \cite{khan2020human,han2018going}, and sport-analysis \cite{zhu2006player,soomro2014action}, to mention just a few. 
Different AR solutions require different data, such as still images \cite{liu2020few} \cite{jeong2021few}, sequences of images \cite{bo2020few} \cite{careaga2019metric}, and sequences of 2D/3D skeleton data \cite{ren2020survey} \cite{wang20213d}. 
The AR-based on sequences of 2D/3D skeleton data has the advantage to use less computation power and ignores the background, focusing only on the human pose \cite{perrett2021temporal}.
To train an AR pipeline, we often need big labeled datasets, especially for Deep-Learning models.
These datasets require manual collection and annotation, and the resulting model requires a fine-tuning, or worse, full retraining to simply add or change a class. These models are badly suited for those applications in robotics in which the robot is required to learn new actions from a few demonstrations and, perhaps more importantly, should be able to recognize when a given input belong to one of the ``known'' classes.
A class of models called Few-Shot Learning (FSL) aims at solving the problem above by adapting to recognize new classes given few samples.
The FSL models for AR are called Few-Shot Action-Recognition (FSAR) models \cite{guo2020attentive, wu2019few}.
Usually, in a few-shot problem, the model receives as input a set of few examples for each class called \emph{support set} and a query; then, the model outputs the probabilities for the query to belong to a class in the support set.
However, in practical applications, it is not possible to foresee in advance all the actions humans will perform while interacting with the robot. In this ``Open Set" (OS) scenario, we cannot add all the possible classes of actions (which are infinite) to the support set and reliably classify each sample.
The OS problem combined with the FSL problem yields the Few-Shot Open-Set Recognition (FSOSR) \cite{liu2020few} problem and its relative application to AR is called Few-Shot Open-Set Action Recognition (FSOSAR).
Typically this problem is solved by creating a ``reject" class, whose instances can be sampled from the training set and can be inferred from the distances between the query and the elements of the support set \cite{liu2020few} \cite{jeong2021few}.
While the FSOSAR problem has been investigated for images \cite{liu2020few} \cite{jeong2021few}, to the best of our knowledge, there is no work that addresses the FSOSAR problem for sequences.
\begin{figure}[t!]
    \centering
\includegraphics[width=1\columnwidth]{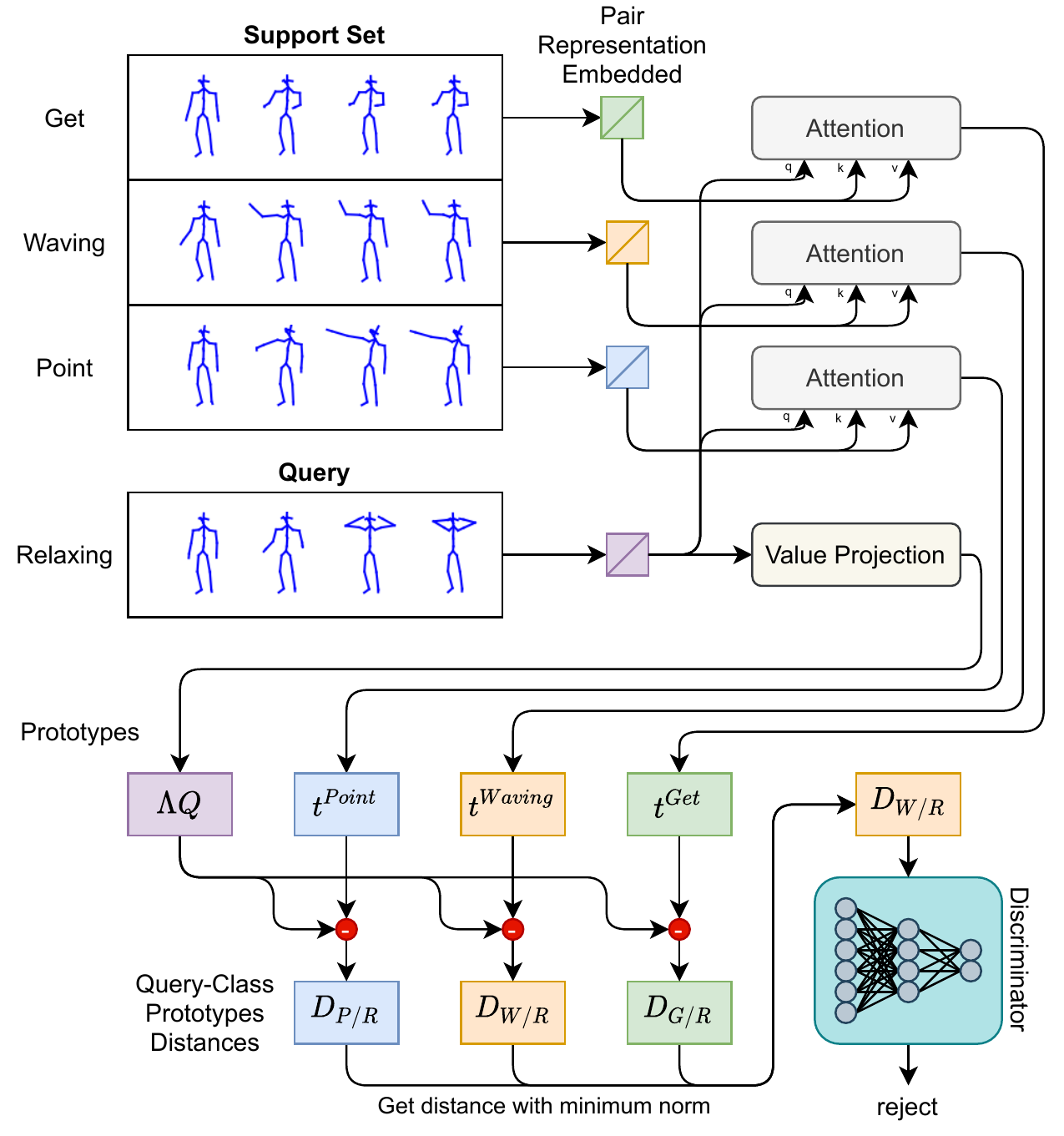}
    \caption{An overview of our approach for the One-Shot Three-Way problem. The query class ''relaxing" is not present in the support set, but the FS part of the model chooses the ''waving" class as the solution. Our discriminator \emph{Disc} discards the FS prediction because it provides a low confidence score (0.11). Thus, our approach selects ''reject" (i.e. ``unknown") as the final class.}
    \label{IN.fig.front}
\end{figure}
In this work, we propose a solution to the FSOSAR problem by defining a discriminator that assigns a confidence score to the difference between the features of the query and the features of the predicted class.
With episodic end-to-end learning, we force this difference to have a high norm value when the difference is high and low otherwise.
Moreover, the discriminator forces this difference to contain information about similarities and differences between the query prototype and the query-class prototype of the predicted class. Finally, we propose to use features obtained from a 3D skeleton of the human, resulting in a simpler model that is less computationally expensive and easier to train.
An overview of our approach is depicted in Figure~\ref{IN.fig.front}.
Our contributions can be summarized as follows:
\begin{itemize}
    \item We created a model that deals with Few-Shot Open-Set learning for sequences of 3D skeleton.
    \item We proposed an episodic end-to-end learning technique that balances the outcome of the discriminator and allows to jointly train the FSL part and the OS part of the model.
    \item We demonstrated qualitatively that the model is capable of differentiating similar but semantically different actions.
\end{itemize}
We released the code\footnote{Available at \url{https://github.com/stefanoberti/ISBFSAR}} and the dataset\footnote{Available at \url{https://bit.ly/3OcD9yb}} in order to make all our experiments easily reproducible.

\section{Related Work}
In this section we give a brief review of various FSL methods, analyzing the advantages and disadvantages of different types of input data and the main differences with respect to our approach.
\subsection{Few-Shot Action Recognition (videos)}
A recent work \cite{careaga2019metric} proposed to aggregate frame-level features of both the RGB image and the optical flow across time.
For the time-aggregation of the features, they tested Pooling, Long-Short Term Memory (LSTM), ConvLSTM, and 3D-Convolutions.
As a FSL algorithm, they tested Matching Networks, Prototypical Networks, and Learned Distance Metrics.
Another work \cite{bo2020few} proposed to use multiple Temporal Attention Vectors to encode video-wide temporal information that adapts to the various lengths of frame sequences. 
The importance scores are then used to select important temporal information.

The works above use sequences of images as input data that allow taking into consideration the whole context, like the objects that the human is interacting with and the surrounding elements of the environment.
In contrast, our work focus on 3D human skeleton data as we do not need to train a Convolutional Neural Network to extract features from the important parts of the RGB images.
As a result, our model requires much less computational power to train and fewer samples when the action is mainly discriminated through the human pose.

\subsection{Few-Shot Action Recognition (skeleton)}
A recent work \cite{wang20213d} tackles the Skeleton-Based FSAR problem with a variant of Dynamic Time Warping. 
It tries to find the best alignment in the temporal and simulated camera viewpoint spaces between the query and the support frames. 
A temporal block encoder expressed the sequences.

Our work also aims at the best temporal alignment between the query and the support set, but in contrast we do not explicitly simulate different camera viewpoints since the embedding layer, trained with a dataset where actions are recorded from different orientations, makes our model implicitly invariant to skeleton rotations.

\subsection{Few Shot Open-Set Recognition (images)}
The FSOSAR problem addresses two problems at the same time: the FSAR problem and the OS problem.
FSOSAR has been studied in the literature by \cite{liu2020few} and \cite{jeong2021few}, but mainly using still images as input data.
Some approaches \cite{liu2020few} propose to solve the FSOSR problem with a combination of Meta-Learning, a random selection of novel classes per episode, a loss that maximizes the posterior entropy for examples of those classes, and a new metric learning formulation based on the Mahalanobis distance \cite{mahalanobis1930test}.
Another approach \cite{jeong2021few} extended the one above, taking inspiration from the concept of reconstruction-based OS recognition methods, where an algorithm is trained to reconstruct the training data and it fails at reconstructing unseen samples.
In this work the authors trained a transformation function based on the fact that the distance between a query feature and its class prototype is closer than the distance between an unknown class feature and that prototype.
Then they compute the distance between the transformed query and the transformed elements of the support set that are not chosen as the class and they check that the difference is under a certain threshold.
In this way, they avoid dependency on pseudo-unseen examples.

In contrast to these works, our approach considers sequences as input data and can therefore discriminate ambiguous frames based on previous ones. For example, by using sequences we can understand if a person is sitting down or standing up, but that is tricky to understand from a single image.

\section{Background}
The Temporal-Relational Cross-Transformer (TRX)\cite{perrett2021temporal} is a few-shot method for action recognition that considers the similarity between all the possible sub-sequences of length $2$ of the query frames and all the possible sub-sequences of length $2$ of all the elements in the support set (the model was also extended to sub-sequences of length ${> 2}$, but in this paper we focus only on pairs of frames).
We are interested in the $K$-way $N$-shot problem with $N=1$ (also known as One-Shot).
With an attention mechanism, the TRX model constructs ``query pair"-specific class prototypes and selects the class relative to the element in the support set which has a minimum query-class difference.
Given a query ${Q \triangleq \{q_1, ..., q_F\}}$, the query representation for the pair $p$ is defined as:
$$Q_p \triangleq [\phi(q_{p_1}) + PE(p_1), \phi(q_{p_2}) + PE(p_2)] \in R^{2 \times D}$$ where $p_1$ and $p_2$ are ordered indices of frames such that ${1 \leq p_1 < p_2 \leq F}$, $F$ is the total number of frames, ${\phi : R^{H\times W\times 3} \rightarrow R^{D}}$ is a convolutional network that extracts a D-dimensional embedding from each input frame and $PE$ is the positional encoding \cite{gehring2017convolutional}.
We define the set of all possible pairs as:
$${\Pi \triangleq \{ (n_1, n_2) \in \mathbb{N}^2: 1 \leq n_1 < n_2 \leq F\}}$$
The support set $S$ contains a support $S^c = \{s^c_1, ..., s^c_F\}$ for each class $c$.
A single frame-pair representation of the video with class $c$ in the support set with respect to the ordered pair of indices ${m = (m_1, m_2)\in \Pi}$ is:
$$S^c_{m} \triangleq [\phi(s^c_{m_1}) + PE(m_1), \phi(s^c_{m_2}) + PE(m_2)] \in R^{2\times D}$$
and the set of all pair representations in the support set for class c is:
$${S^c \triangleq \{ S^c_{m} : (m \in \Pi)\}}$$
By using three different linear projections called queries ${\Upsilon: R^{D\times 2} \rightarrow R^{D_\Upsilon}}$, with  key ${\Gamma: R^{D\times 2} \rightarrow R^{D_\Gamma}}$ and value ${\Lambda: R^{D\times 2} \rightarrow R^{D_\Lambda}}$, we can compute the query-specific class prototype for the query $Q_p$ with respect to the class $c$:
$${t_p^c \triangleq Softmax(L(\Gamma \cdot S_{m}^c) \cdot L(\Upsilon \cdot Q_p)) \cdot (\Lambda \cdot S_{m}^c)}$$
where ${t_p^c \in R^{D_V}}$ and $L$ is the standard layer normalisation \cite{ba2016layer}.
Then we compute the distance between the query $Q_p$ mapped with  $\Lambda$ and each query-specific class prototype of the support set $S^c$:
$${T(Q, S^c) \triangleq \frac{1}{|\Pi|}\sum_{p \in \Pi}||t^c_p - \Lambda \cdot Q_p||}$$
In this way, we can compute a distance between the query and each class of the support set.
The model is then trained with a cross-entropy loss over the query-class negative distances $T(Q, S^c)$.
\section{Problem Formulation}
\label{PF}
In this section, we formally formulate the problem of FSL and we extend it to the FSOSR. 
We pose the problem as a single-instance classification problem, where each class encodes a different action.
Let ${D^S \triangleq \{x_i^S, y_i^S\}_{i=1}^{NK}}$ be the support set, where $x_i^S$ and $y_i^S$ represent an instance and its label respectively, $K$  the number of classes, and $N$  the number of labeled samples for each class.
Given a known query set ${{D^{Kn} \triangleq \{x_i^{Kn} \in X^{Kn}, y_i^{Kn} \in Y^S\}_{i=1}^{Q}}}$, where $Q$ is the number of query samples for each class, the aim of FSL is to infer for each query sample $x_i^{Kn}$ the corresponding class $y_i^{Kn}$ that is assumed to always be contained inside the support labels $Y^S$.

In an FSOSR problem, the query set contains also instances of classes that are not part of the support set.
The query set ${D^Q \triangleq D^{Kn} \cup D^{Un}}$ equals the union of both the known query set ${D^{Kn}}$ and the unknown query set ${D^{Un} \triangleq \{x_i^{Un} \in X^{Un}, y_i^{Un} \in Y^{Un}\}_{i=1}^{N^{Un}}}$, where ${Y^S \cap Y^U = \emptyset}$ holds and $N^{Un}$ is the number of unknown instances.
Thus, a FSOSR problem requires to create a classifier $f$ such that given a support set $S$, for each query sample ${x_i}$ and true class ${y_i}$, the following holds:
\begin{equation}
\label{eq:desired-f}
\centering
    \bar f(x_i)= 
\begin{cases}
    y_i,& \text{if } y_i \in Y^S\\
    False,              & \text{otherwise}
\end{cases}
\end{equation}

Since previous work focuses on the FSOSR problem for still images \cite{liu2020few} \cite{jeong2021few} and fail to correctly discriminate some actions as for ``standing-up" and ``sitting-down", we address the problem using sequences of 3D skeletons.
We focus on sequences of 3D skeletons because we are mostly interested in the human pose, we want to be able to differentiate temporal actions, and because working with skeleton data requires much less computational power (at least for the training).
\section{Proposed Solution}
In this section, we present a novel model for FSOSR with its relative end-to-end training technique that allows to increasingly train the discriminator with a balanced number of positive and negative samples.
An overview of our proposed solution is depicted in Figure~\ref{IN.fig.front}.
\subsection{Model}
Our proposed model is a variation of the TRX \cite{perrett2021temporal} model.
To adapt it for sequences of skeleton, we replaced the convolutional neural network ${\phi : R^{H\times W\times 3} \rightarrow R^{D}}$ with an \emph{embedding} network ${\Psi : R^{J\times 3} \rightarrow R^{D}}$ where $J$ is the number of joints of the skeleton.
We used a Multi-Layer Perceptron \cite{kubat1999neural} as an embedding network, with a hidden layer of size $J\times 3\times 2$ and an output layer of size $D$.
We also added a ReLU \cite{agarap2018deep} activation function after each layers.
Given different skeleton rotations during training, the embedding function $\Psi$ learns to project similar poses with different rotations into two close $D$-dimensional vectors.

To extend to the OS this FSAR model, we need to add a ``reject" class that indicates that the query class $y_i$ is not contained inside the support set classes $Y^S$.
We define ${t^c \triangleq stack(\{t_p^c: p \in \Pi\})}$ and ${\tilde{q} \triangleq stack(\{\Lambda \cdot Q_p: p \in \Pi \})}$, where $stack$ is a function that concatenates the elements a sequence along a new dimension, such that ${t^c \in \mathbb{R}^{|\Pi| \times  D^\Lambda}}$ represent the concatenation of the pairs of query-class prototypes and ${\tilde{q} \in \mathbb{R}^{|\Pi| \times  D^\Lambda}}$ represents the concatenation of the pairs of query prototypes.
We could extend to the OS by comparing the query-class prototypes $t^c$ of each class $c$ with the query prototypes $\tilde{q}$, but this would lead to many negative samples and an unbalanced training of the discriminator.
Instead, we propose an alternative approach in which an action is associated to a set of known actions and a discriminator is trained to accept or reject the best association. 
We add this behaviour to our model in the following way: first we get the class $c_{FS}$ that is chosen by the FSL part of the model 
$$c_{FS} \triangleq \underset{c}{\operatorname{argmin}} \{T(Q, S^c)\}$$
, then we give the matrix $\tilde{q} - t^{c_{FS}}$ to a discriminator $Disc$.
The discriminator $Disc$ assigns a score $[0, 1]$ to its input that represents the confidence of associating the query-class features difference with two semantically close skeleton sequences.
We implemented $Disc$ with a Multi-Layer Perceptron as follows: the first layer $l^1: \mathbb{R}^{D^\Lambda} \rightarrow D^r$ reduces the dimension of each element of the sequence, obtaining a matrix with size $\mathbb{R}^{|\Pi| \times  D^r}$.
Then it flattens the last two-dimension and uses three layers to progressively reduce the dimension of the array to one neuron, with a ReLU \cite{agarap2018deep} function between each pair of layers.
Finally, we apply a Sigmoid function to obtain a score in $[0, 1]$ that we compare with a threshold value to either ''accept" or ''reject" the class predicted by the FSL part of the model.
Thus, the output of our model holds the following:
\begin{equation}
\label{eq:proposed-f}
    f(x_i, S)= 
\begin{cases}
    c_{FS},& \text{if } Disc(\tilde{q} - t^{c_{FS}}) > \tau\\
    False,              & \text{otherwise}
\end{cases}
\end{equation}

where $\tau$ is a threshold that depends on the similarity of the actions we are considering.
Basically, if the actions in the support set are very similar, a higher threshold value $\tau$ may be needed to correctly reject unknown queries.
\subsection{Training}
The model is trained end-to-end with episodic learning.
To train the FSAR part of the model, we follow the approach of TRX \cite{perrett2021temporal}: given a query $Q_i = \{q_{i1}, ..., q_{iF}\}$ with true class $y_i \in Y^S$ and a support set $S_i = \{S_i^1, ..., S_i^k\}$ with $K$ examples (one for each class $c$), we compute the FS loss with a cross-entropy between the query-class negative distances and a one-hot target vector for the task $i$ is:
$$\ell_{FS} \triangleq \frac{1}{|B|}\sum_{i=0}^B CE(stack(\{-T(Q_i, S_i^c): c\in Y^S\}), \tilde{y}_i)$$
where $B$ is the number of training samples, $CE$ stands for Cross-Entropy and $\tilde{y}_i$ is an one-hot vector with the entry corresponding to $y_i$ set to $1$.

To jointly train the discriminator, we need to feed it with positive and negative samples.
To train the model end-to-end we need to consider only true positive samples, that are samples $Q_i$ that were correctly classified by the FSAR part of the model, and so that $c_{FS, i} = y_i$.
We define $z$ as the number of samples correctly classified in a minibatch by the FSAR part of the model.
Now that we have $z$ positive samples, we can obtain the negative samples by just sampling queries $Q_i$ with label $y_i$ from the unknown set $D^U$.
By sampling $z$ negative samples we can keep the training of the generator balanced, where the number of samples that it receives at each training step is ${z \times 2}$ and it increases during training accordingly to the accuracy of the FSAR part of the model.
This also allows us to backpropagate the OS loss to the FSAR part of the model only where the query and the most similar element in the support set are similar, providing more information to the training.
We define this loss as:
$$\ell_{OS} \triangleq \frac{1}{|B|}\sum_{i=0}^B\tilde{\ell}_{OS}(Q_i, S_i^c)$$
where, once defined $\tilde{o}_i \triangleq Disc(\tilde{q}_i-t_i^c)$:
\[
    \tilde{\ell}_{OS}(Q_i, S_i^c)= 
\begin{cases}
    BCE(\tilde{o}_i,1),& \text{if } c_{FS, i} = y_i\\
    BCE(\tilde{o}_i,0),& \text{if } y_i \notin Y^S\\
    0,& \text{if } y_i \in Y_i^S \wedge c_{FS,i} \neq y_i
\end{cases}
\]
where BCE stands for Binary Cross-Entropy.
By combining the FS loss and the OS loss we obtain the total loss:
$${\ell_{FSOS} \triangleq \ell_{FS} + \sigma \ell_{OS}}$$

where $\sigma$ is a regularizer value.

\section{Experimental Validation}
In this section, we provide the experimental validation of our model. 
In particular, we investigate the performance of our model in recognizing new classes from a small set of examples and the ability of the model to detect and discard ``unkown'' actions.
We compare our model with the baseline, providing quantitative and qualitative results.
We then show that, for the classes in the test set, our approach solves the problem defined in Section~\ref{PF}.
The experimental validation characterizes in:
\begin{itemize}
    \item Quantitative validation: we compared the accuracy of the FSOS task of our model with respect to a baseline, where the discriminator is not present and a query is ``rejected" if the exponential of the negative query-class distance is lower than a certain threshold.
   \item Qualitative validation: we discussed some predictions of our model over the test set, demonstrating its strength to avoid recognizing wrong actions.
\end{itemize}
Unfortunately, to the best of our knowledge, there are no other works in the literature that studied the FSOSAR problem for sequences of 3D skeleton and therefore we cannot make a fair comparisons with other methods on this task.
Thus, instead, we created a baseline where the discriminator is removed and the OS loss is computed directly on the negative query-class prototypes distance without further processing.
Since the baseline does not use a discriminator explicitly trained on the OS task, we expect it to have lower accuracy.
\subsection{Comparison with Baseline (Quantitative validation)}
We declare a simple baseline called \emph{EXP} where the discriminator is removed and the confidence is computed with an exponential function over the maximum negative distance $\max_c\{-T(Q, S^c):c\in Y^S\}$.
Indeed, since the query-class negative distances are bounded in $(-\infty, 0]$, its exponential has value inside $[0, 1]$ that can be used as a confidence score to either confirm or ``reject" the class predicted by the FSL part of the model.

As dataset, we consider the NTURGBD-120 dataset \cite{liu2019ntu}.
This dataset contains 120 classes involving daily actions, medical actions, and two people interactions.
Each class has about a thousand samples with relative RGB videos, depth map sequences, 3D skeletal data, and infrared (IR) videos.
However, some samples do not have skeleton data and, as pointed out also in the literature \cite{trivedi2021ntu}, most of the given skeletons have low quality and are noisy.
Because of this, we decided to re-extract the skeleton from the RGB videos.
Thus, we took $16$ equidistant frames from each video, we extracted the 3D skeleton with MetrABS \cite{sarandi2020metrabs} and we centered the skeleton with respect to the pelvis coordinate.
The output of the MetrABS heads is already bounded inside $[-1, 1]$, and therefore it did not need normalization. On the other hand,  we decided not to normalize the rotation because we want the embedding layer to be invariant to the rotation of the skeletons.
We removed the 26 classes representing two people interactions since our problem focuses on a single skeleton at a time.
To evaluate the performance of our method, we use an approach similar to the One-Shot Action Recognition described in the literature  \cite{liu2019ntu}, but adapted for the OS task: we randomly select $k$ classes and for each of them we add in the support set the corresponding exemplar \cite{liu2019ntu} (one fixed action instance for class), then we compute the metric on all the elements of the test set.
In this way, we have a known set $D^K$ that contains all the samples of the $k$ selected class and an unknown set $D^U$ that contains all the samples from the remaining classes.
We compute a metric that we call \emph{FSOS-ACC}, which is the accuracy between the predicted classes and the target classes, considering also the ``reject" class.
We expect our model to classify as ``reject" all the instances in the unknown set $D^Un$ and to classify with the correct classes in the support set all the instances of the known set $D^Kn$.
We performed the comparisons for growing values of $K$; for each $K$ we train the model $100$ times and we report the mean accuracy and its standard deviation.

\begin{figure}[t!]
    \centering
\includegraphics[width=1\columnwidth]{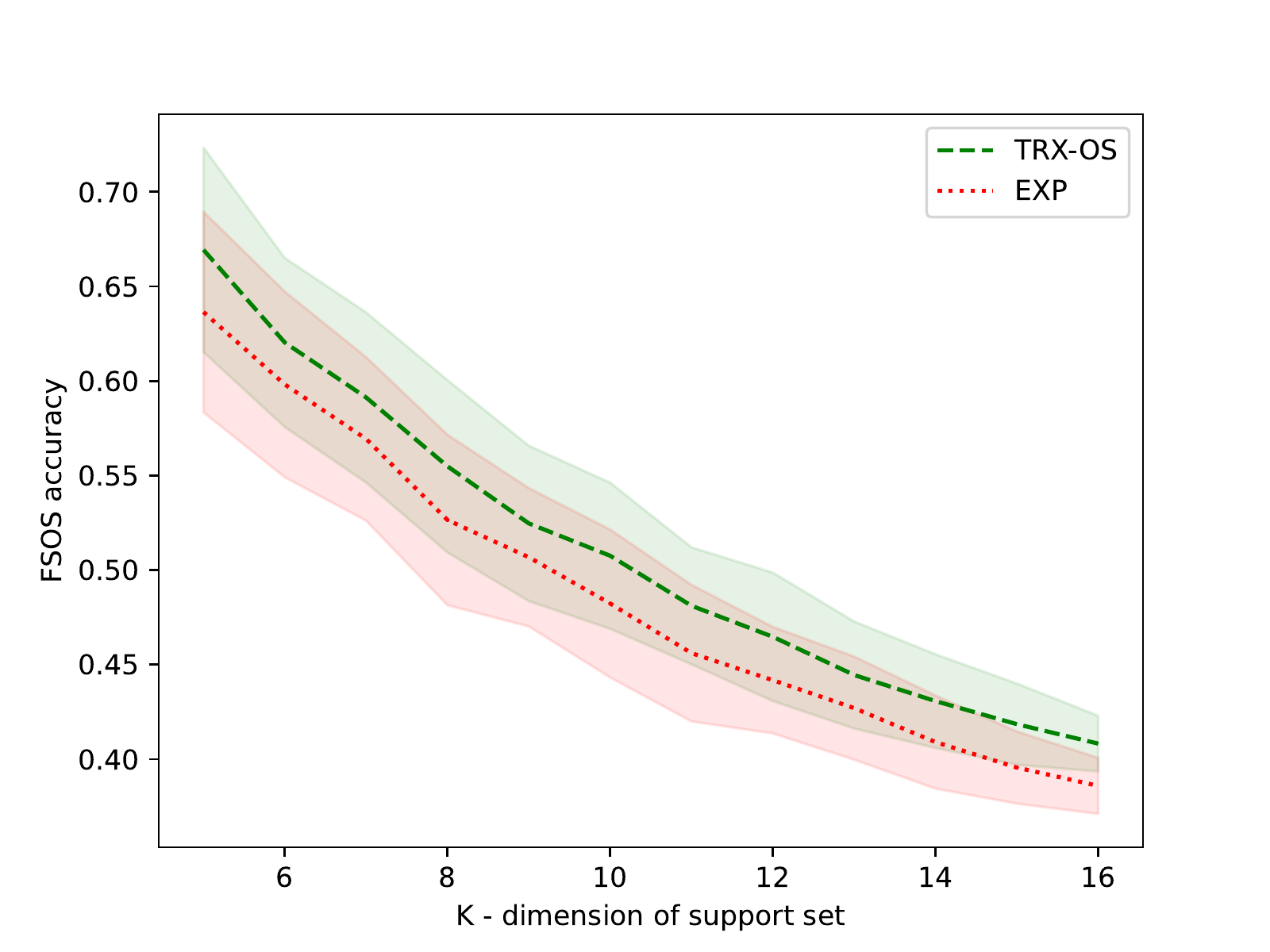}
    \caption{Comparision between the proposed model (\emph{TRX-OS}) and the baseline (\emph{EXP}) by the FSOS-ACC metric on our dataset. The \emph{TRX-OS} model outperforms the baseline for every $K$.}
    \label{res_fsos}
\end{figure}

\begin{table}
\centering
\begin{tabular}{ |c|c|c|c|c| } 
 \hline
  & K=5 & K=10 & K=15\\ 
 \hline
 TRX-OS & \textbf{0.67$\pm$ 0.05} & \textbf{0.52$\pm$ 0.04} & \textbf{0.43$\pm$ 0.02} \\ 
 EXP  & 0.63$\pm$ 0.05 & 0.51$\pm$ 0.04 & 0.41$\pm$ 0.02 \\ 
 RANDOM  & 0.17        & 0.09           & 0.06 \\
 \hline
\end{tabular}
\caption{FSOS-ACC metric for our model (TRX-OS) and the baseline (EXP). We also show random results for reference.}
\label{comparision}
\end{table}

Figure~\ref{res_fsos} depicts the results of our comparison, while Table~\ref{comparision} shows the numerical values.
For higher values of $k$, there is an increasing number of instances that should be rejected by the discriminator.
Moreover, the more classes there are in the support set, the harder is to match the query with the true class because it is more probable that some classes are similar.
With $k=5$ we can reject or select the true class the $67\%$ of the times, that means that our model can be reliably used in real Human-Robot scenarios when the number of classes is limited.
Our model outperforms the baseline for all the values of $k$.
This means that the discriminator learns useful concepts that are also backpropagated to the FSL part of the model thanks to the Open-Set loss $\ell_{OS}$.

\subsection{Discriminator's Confusion Matrix (Quantitative validation)}
Another quantitative analysis refers to the Open-Set Accuracy of our model.
In fact, if we set our support set size to one we can build a confusion matrix on the output of the discriminator to get insights on which actions are more difficult to classify.
We use the exemplar of one class from the test set as the support set, then we attempts to classify all the elements of the test set and we computed the number of instances that are classified as $True$ by our model for each class.
We show a subset of the results of this analysis in Table ~\ref{confmatr}.
We expect the elements on the diagonal to have a value close to one and all the other elements of the matrix to have a value close to zero.
Our model performs well in this task for some classes like \emph{Take off glasses} and \emph{Falling}, but other classes like \emph{Tear up paper} and \emph{Apply cream on face} have low performances.
This happens mainly for two reasons, as discussed also in the following Subsection ~\ref{qual}: 
\begin{itemize}
    \item The skeleton model that we used for training and testing does not track the hands of the human. As a consequence, it is impossible to differentiate between actions like \emph{Apply cream on face} and \emph{Hush (quite gesture)} because the focus of these actions is on the hand movement and the fingers position.
    \item By using skeleton data only, we ignore which object the human is interacting with. Due to this fact, actions like \emph{Stample book} and \emph{Tear up paper} are very similar.
\end{itemize}
An important aspect of this analysis is that we set the threshold value $\tau$ of Equation~\ref{eq:proposed-f} to $0.5$, but different values could be more suitable for different applications.

\subsection{Qualitative results}
\label{qual}
In our applicatio,n when the robot detects an action performed by the human it triggers his motors and assists the human in the completion of the action.
Since these actions involves also dangerous Human-Robot Interactions (e.g. the robot passes a box to the human), it is needed that the robot does not misclassify critical human actions.
In the following, we show and discuss some qualitative examples that justify the existence of our model.
We show three examples in the 3-Way 1-Shot scenario for ease of representation.

\begin{table}[t!]
\centering
\begin{tabular}{ |c|cccccccc| } 
 \hline
  & \rot{Take off glasses } & \rot{Tear up paper} & \rot{Pointing something } & \rot{Falling} & \rot{Hush (quite) } & \rot{Stample book } & \rot{Apply cream on face } & \rot{Arm circles }\\ 
 \hline
  &&&&&&&&\\
 Take & 0.91 & 0.01 & 0.03 & 0. & 0.01 & 0. & 0.05 & 0.\\
 &&&&&&&&\\
 Tear & 0. & 0.42 & 0.02 & 0. & 0. & 0.49 & 0.06 & 0.\\
  &&&&&&&&\\
 Pointing & 0. & 0.03 & 0.82 & 0. & 0. & 0. & 0. & 0.\\
  &&&&&&&&\\
 Falling & 0. & 0. & 0. & 0.87 & 0. & 0. & 0. & 0.\\
  &&&&&&&&\\
 Hush  & 0.09 & 0. & 0.02 & 0. & 0.74 & 0. & 0.48 & 0.\\
  &&&&&&&&\\
 Stample  & 0. & 0.06 & 0. & 0. & 0. & 0.52 & 0. & 0.\\
  &&&&&&&&\\
 Apply & 0.12 & 0. & 0.02 & 0. & 0.48 & 0. & 0.6 & 0.\\
  &&&&&&&&\\
 Arm & 0. & 0.03 & 0.02 & 0. & 0. & 0. & 0. & 0.71\\
  &&&&&&&&\\
 \hline
\end{tabular}

\caption{Confusion matrix of our model for the OS task with respect to the test classes. On the rows, we have the classes that were added in the support set, and on the columns, we have the classes that were tested against the exemplar for the class in the support set.}
\label{confmatr}
\end{table}

\begin{figure}
    \centering
\includegraphics[width=1\columnwidth]{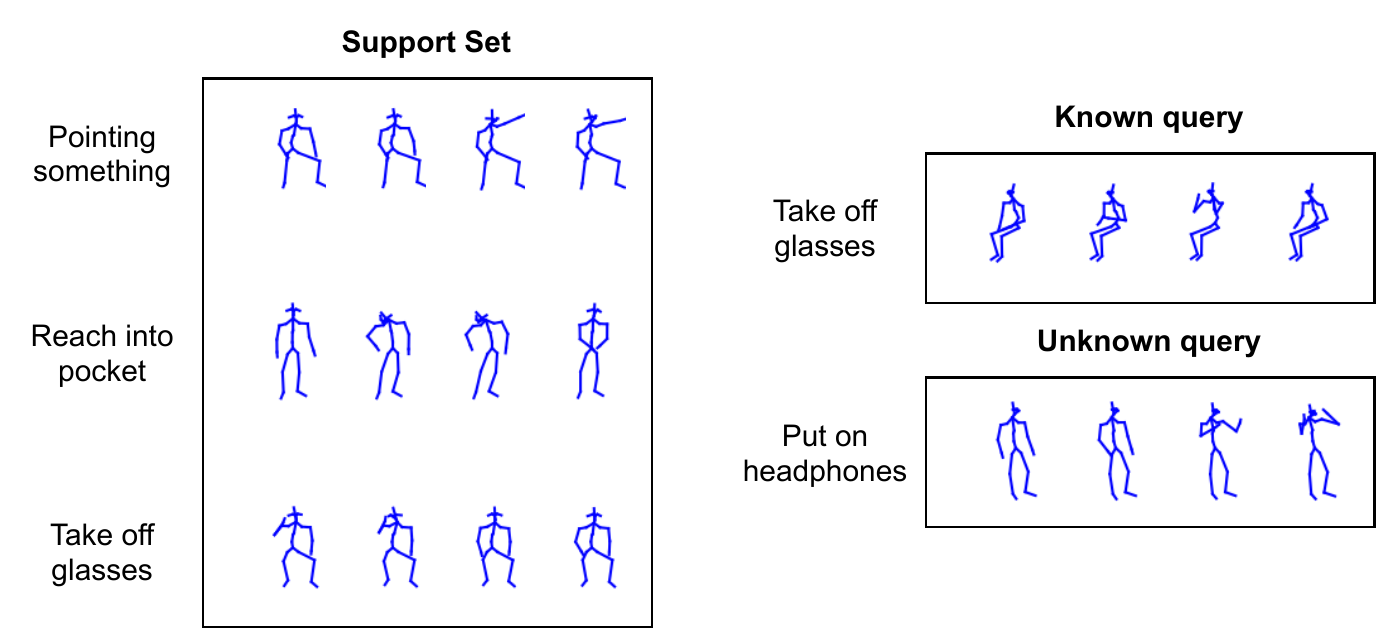}
    \caption{First qualitative example visualization. On the left the support set and on the right the known query and the unknown query.}
    \label{es1}
\end{figure}

\begin{table}[t!]
\centering
\begin{tabular}{ |c|c|c| } 
 \hline
  & Take (know) & Put (unknown)\\ 
 \hline
 Point      & 0.   & 0.24\\ 
 Reach      & 0.   & 0.04\\ 
 Take       & 0.99 & 0.72\\ 
  \hline
 OS-Score   & 0.87 & 0.09\\ 
 \hline
\end{tabular}
\caption{First qualitative example results. The first three rows shows the FSL scores for the known/unknown query and all the elements in the support set, the last row shows the discriminator's confidence between the known/unknown query and the element of the support set with highest FSL score.}
\label{es1tab}
\end{table}

In the first example, depicted in Figure~\ref{es1} with results shown in Table~\ref{es1tab}, both the FS and OS parts of the model correctly classify the known query \emph{Take off glasses}, showing that the model effectively learned to deal with different rotations of the skeleton.
The unknown query \emph{Put on headphones} is similar to the exemplar of the class \emph{Take off glasses} in the support set, but in the first action, the object is brought closer to the head while in the latter the object is moved away from the head.
Indeed, the FS part of the model chooses the \emph{Take off Glasses} action because it is the most similar in the support set, but the discriminator correctly rejects the unknown query, showing that the discriminator also learned to deal with temporal information.

\begin{figure}[t!]
    \centering
\includegraphics[width=1\columnwidth]{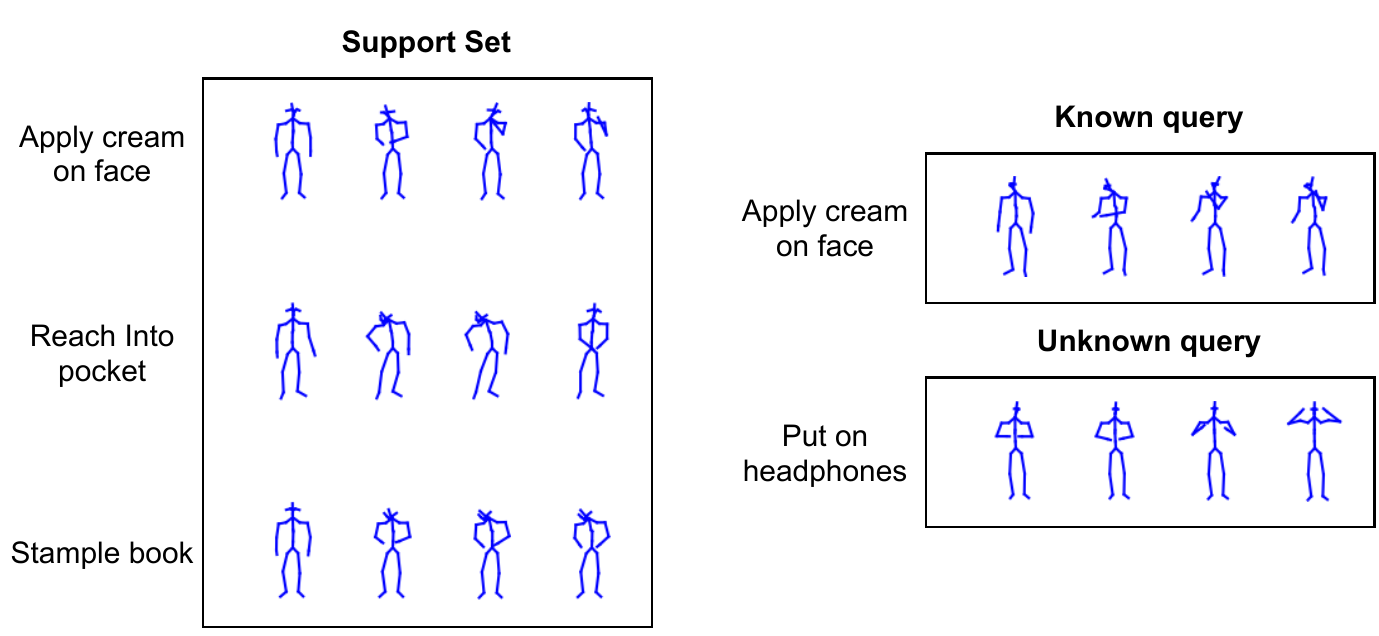}
    \caption{Second qualitative example visualization. On the left the support set and on the right the known query and the unknown query.}
    \label{es2}
\end{figure}

In the second example, depicted in Figure~\ref{es2} with results shown in Table~\ref{es2tab}, both the FS and OS parts of the model correctly classify the given known query \emph{Apply cream on face} as one class in the support set, with a high OS-Score. 
This case is not too difficult for the FS part of the model because the \emph{Apply cream on face} action is the only one of the support set where the arms reach the head's height.
For the same reason, the FS part of the model classifies the known query \emph{Put on headphones} with high confidence as \emph{Apply cream on face} function, but then the discriminator rejects the prediction.

\begin{table}[t!]
\centering
\begin{tabular}{ |c|c|c| } 
 \hline
  & Apply (known) & Put (unknown)\\ 
 \hline
 Apply      & 0.99   & 0.98\\ 
 Reach      & 0.0   & 0.1\\
 Stample       & 0.   & 0.1\\ 
  \hline
 OS-Score   & 0.98 & 0.31\\ 
 \hline
\end{tabular}
\caption{Second qualitative example results. The first three rows shows the FSL scores for the known/unknown query and all the elements in the support set, the last row shows the discriminator's confidence between the known/unknown query and the element of the support set with highest FSL score.}
\label{es2tab}
\end{table}

\begin{figure}[t!]
    \centering
\includegraphics[width=1\columnwidth]{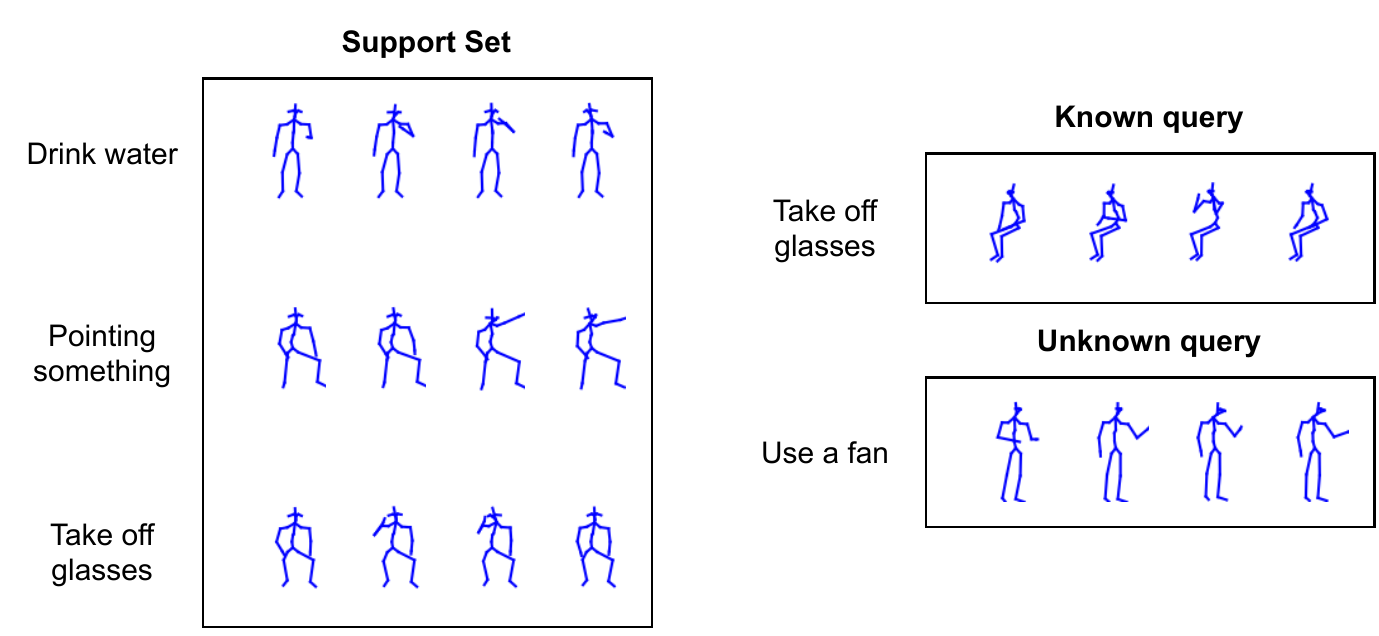}
    \caption{Third qualitative example visualization. On the left the support set and on the right the known query and the unknown query.}
    \label{es3}
\end{figure}

\begin{table}[t!]
\centering
\begin{tabular}{ |c|c|c| } 
 \hline
  & Take (known) & Use (unknown)\\ 
 \hline
 Drink       & 0.   & 0.02\\ 
 Point      & 0.   & 0.97\\ 
 Take      & 0.99   & 0.01\\ 
  \hline
 OS-Score   & 0.87 & 0.80\\ 
 \hline
\end{tabular}
\caption{Third qualitative example results. The first three rows shows the FSL scores for the known/unknown query and all the elements in the support set, the last row shows the discriminator's confidence between the known/unknown query and the element of the support set with highest FSL score.}
\label{es3tab}
\end{table}

The third example, depicted in Figure~ \ref{es3} with results shown in Table~\ref{es3tab}, is interesting because it highlights the limitation of the 3D skeleton data.
As in the other examples, the FSL part of the model chooses the right action in the support set for the known query \emph{Take off glasses} and the discriminator confirms the prediction.
Anyway, the unknown query tricks our model.
Indeed, the unknown action \emph{Use a fan} is performed only with wrist movements and does not involve movements from the rest of the arm.
Since MetrABS does not track hand movements, the sequence of skeletons extracted from the video does not describe correctly the action and, just by looking at them, it is easy to think that the human is extending his arm towards something and nothing else.
Because of this, the FS part of the model chooses \emph{Pointing something} as the candidate class and the OS part of the model confirms the wrong prediction since it does not know about the hand movements.

\begin{figure}[t!]
    \centering
\includegraphics[width=0.9\columnwidth]{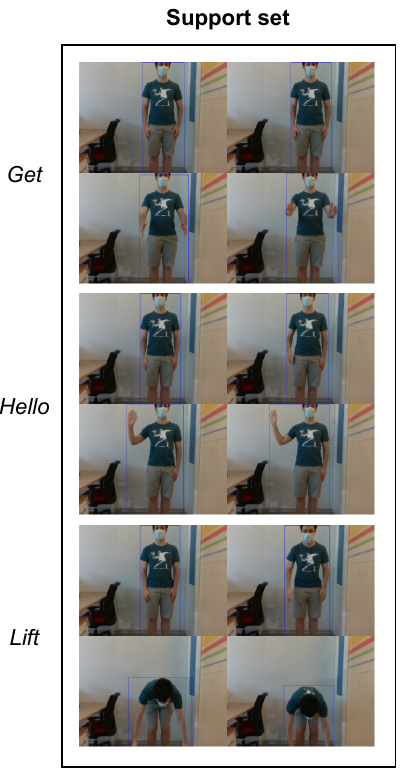}
    \caption{Support set as a reference for the two unknown queries.}
    \label{real}
\end{figure}

\begin{figure}
    \centering
\includegraphics[width=0.7\columnwidth]{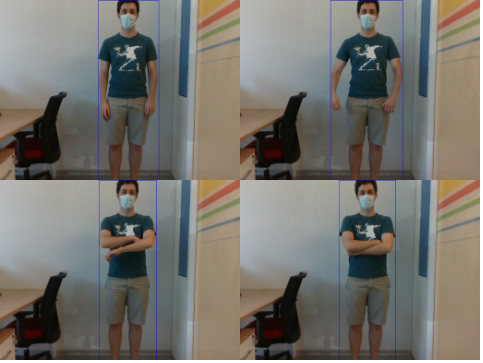}
    \caption{First unknown query: the FS part of the model classifies this query as ''get" with $54\%$ of confidence, but the query is rejected because the Discriminator's output is low (0.07).}
    \label{test1}
\end{figure}

\begin{figure}
    \centering
\includegraphics[width=0.7\columnwidth]{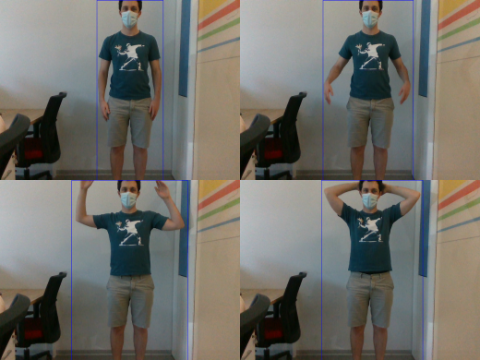}
    \caption{Second unknown query: the FS part of the model classifies this query as ''get" with $99\%$ of confidence, but the query is rejected because the Discriminator's output is low (0.01).}
    \label{test2}
\end{figure}

In Figure~\ref{real} we provide examples with data streamed from a camera. 
The skeleton are extracted from a input video with MetrABS~\cite{sarandi2020metrabs}. 
In our case, we wanted the robot to detect when the human is ready to receive a box (i.e. action ``get"), but reject all those cases in which the human is doing different movements.
We can see that the desired behavior is obtained for both the two queries in Figure~ \ref{test1} and Figure~ \ref{test2}: the model associates the actions as closer to ``get", however, the decision is rejected based on the the OS score, that is low for both examples.
This demonstrates that, without the discriminator and the OS score, the system would classify all the unknown queries as actions ``Get", producing unwanted robot behavior.


We showed that our model can efficiently deal with the FSOS problem, avoiding to incorrect classifying queries with the true class that does not belong to the support set.
Thus, for the classes tested $x_i$ and the support set $S$ out proposed approach, encoded by $f(x_i, S)$ (in Equation \eqref{eq:proposed-f}) approximates the function $\bar f(x_i)$ (in Equation \eqref{eq:desired-f})  of problem defined in Section~\ref{PF}, and outperforms the baseline.

\section{Conclusion}
In this paper we introduced a novel model to deal with the Few-Shot Open-Set Recognition problem for sequences of 3D skeleton data.
A discriminator is added to confirm or ``reject" the difference between the query prototype and the query-class prototypes of the most similar element in the support set.
We explained how to train the discriminator jointly with the FSL part of the model in such a way to keep the training of the discriminator balanced.
Moreover, the discriminator is trained proportionally with the accuracy of the FSAR part of the model.
We performed a quantitative analysis with a baseline to demonstrate the importance of the discriminator and qualitative analysis, demonstrating that our model can differentiate similar skeleton sequences that perform different actions.
\section{Acknowledgment}
The paper was supported by the Italian National Institute for Insurance against Accidents at Work (INAIL) ergoCub Project.

\balance
\bibliographystyle{unsrt} 
\bibliography{refs} 
\end{document}